\documentclass[a4paper,conference]{IEEEtran}
\IEEEoverridecommandlockouts
\usepackage{cite}
\usepackage{amsmath,amssymb,amsfonts}
\usepackage{algorithmic}
\usepackage{graphicx}
\usepackage{textcomp}
\usepackage{xcolor}
\usepackage{balance}
\usepackage[none]{hyphenat}
\usepackage[numbers]{natbib}
\usepackage{dsfont}
\usepackage{subcaption} 
\graphicspath{{figures/}} 
\usepackage{hyperref}

\def\BibTeX{{\rm B\kern-.05em{\sc i\kern-.025em b}\kern-.08em
    T\kern-.1667em\lower.7ex\hbox{E}\kern-.125emX}}
    
\newtheorem{hypothesis}{Hypothesis}

\begin{document}

\title{Verifying the Causes of Adversarial Examples}

\author{\IEEEauthorblockN{
Honglin Li\IEEEauthorrefmark{1}\IEEEauthorrefmark{4},
Yifei Fan\IEEEauthorrefmark{2},
Frieder Ganz\IEEEauthorrefmark{3},
Anthony Yezzi\IEEEauthorrefmark{2}, and 
Payam Barnaghi\IEEEauthorrefmark{1}\IEEEauthorrefmark{4}}
\IEEEauthorblockA{\IEEEauthorrefmark{1} Department of Brain Sciences,
Imperial College London, W12 0NN, London, United Kingdom. \\
}
\IEEEauthorblockA{\IEEEauthorrefmark{2}School of Electrical and Computer Engineering\\
Georgia Institute of Technology,
Atlanta, Georgia 30308, USA\\ 
}
\IEEEauthorblockA{\IEEEauthorrefmark{3}Adobe,
Grosse Elbstrasse 27, 22767 Hamburg, Germany\\
}
\IEEEauthorblockA{\IEEEauthorrefmark{4}
Care Research and Technology Centre, The UK Dementia Research Institute (UK DRI)
}
}

\maketitle

\begin{abstract}
The robustness of neural networks is challenged by adversarial examples that contain almost imperceptible perturbations to inputs which mislead a classifier to incorrect outputs in high confidence. Limited by the extreme difficulty in examining a high-dimensional image space thoroughly, research on explaining and justifying the causes of adversarial examples falls behind studies on attacks and defenses. 
In this paper, we present a collection of potential causes of adversarial examples and verify (or partially verify) them through carefully-designed controlled experiments. 
The major causes of adversarial examples include model linearity, one-sum constraint, and geometry of the categories. To control the effect of those causes, multiple techniques are applied such as $L_2$ normalization, replacement of loss functions, construction of reference datasets, and novel models using multi-layer perceptron probabilistic neural networks (MLP-PNN) and density estimation (DE). 
Our experiment results show that geometric factors tend to be more direct causes and statistical factors magnify the phenomenon, especially for assigning high prediction confidence. 
We believe this paper will inspire more studies to rigorously investigate the root causes of adversarial examples, which in turn provide useful guidance on designing more robust models. 

\end{abstract}


\section{Introduction} \label{sec:intro}
The past decade has witnessed a tremendous success on machine learning with deep neural networks, especially its application in computer-vision problems. Adversarial examples, however, remain a critical issue which hinders the industry from building robust real-world applications. First discovered in \cite{szegedy2013intriguing}, adversarial examples contain almost imperceptible perturbations to the original inputs which can mislead a classifier to an incorrect output, even in high confidence. 
Since the discovery of such an intriguing property, researchers have been actively studying the topic by proposing algorithms that either attack or defend machine-learning methods. In contrast, studies on revealing, explaining and validating the causes of adversarial examples are far less than those that focus on attackers and defenders. 

Although the research community would agree that studying the cause of adversarial examples is essentially important, one decisive difficulty for such a type of studies is that justifying a relevant hypothesis or statement may often require a thorough examination of the entire proximity of an input sample in a high-dimensional image space. As an efficient ``telescope'' is not yet available for fully observing the geometry of the high-dimensional image universe, the studies are constrained by the limitation of computation resources. 
Fortunately, researchers have proposed thoughtful strategies and designs to conduct empirical studies \cite{fawzi2018empirical, mickisch2020understanding}, which reveal the characteristics of the learning process and potential reasons for the existence of adversarial examples. 
Following a similar methodology, in this paper, we design controlled experiments that help verify the causes of adversarial examples. 

The major contribution of this paper, therefore, lies in the verification of several hypotheses regarding the causes of adversarial examples through carefully-designed controlled experiments. 
The review and collection of explanations and hypotheses on adversarial examples may also become valuable for future reference. 
The paper itself, however, does not contain any adversarial attacks or defenses which we wish to be considered as the state-of-the-art (or tentatively out-of-date in the future). Instead, the hypotheses and their verification process could provide useful references for subsequent studies on both the analysis of the causes and the design of robust machine-learning approaches.  

The rest of the paper is organized as follows. Section \ref{sec:exisiting_explanations} introduces popular explanations on the reasons for adversarial examples. Each subsection in Section \ref{sec:hypotheses_verification} presents a hypothesis on the cause of adversarial examples and its verification process. Section \ref{sec:support} provides important background knowledge which guides the experiment design in the verification process. Finally, Section \ref{sec:conclusion} concludes with summary and discussions. All relevant materials, including codes, data, and pre-trained models are available for download.\footnote{\url{https://github.com/mozzielol/ar}}  

\section{Popular Explanations} \label{sec:exisiting_explanations}
In this section, we review existing explanations on the causes of adversarial examples. Due to the extreme difficulty that often occurs for fully justifying (or even designing experiments for) a hypothesis in this domain, researchers may tend to raise their views and understandings outside of the major contributions (e.g., in the discussion section of their paper) to avoid potential criticism. Consequently, it is possible that some of the boldest yet most potential guesses might become buried in literature and thus not included in this section. In our opinion, however, those views are still valuable even though they may not be fully justified at the moment.

\emph{Low-probability “pockets” in the manifold}: At the time of first discovery, the authors of \cite{szegedy2013intriguing} interpret adversarial examples as ``blind spots'' which belong to low-probability ``pockets'' in the data manifold. To further illustrate the situation, the authors analogize the input domain to the set of rational numbers, which is not dense from a topological perspective. 

\emph{Linearity of the model}: One of the most widely accepted reasons for adversarial examples is the linearity of the model in high-dimensional spaces \cite{goodfellow2014explaining}. The simple and clean explanation states that $w^T(x + \Delta x)$ can differ significantly from $w^Tx$, especially when the dimension of inputs $x$ is quite large for images. In Section \ref{subsec:linearity}, we provide additional supportive evidence on this explanation with results from our controlled experiments. 

\emph{Test error in additive noise}: After observing the error rates in randomly corrupted image distributions, the authors of \cite{gilmer2019adversarial} argue that it should not be surprising to find adversarial examples. They also suggest that improving adversarial robustness should be aligned with improving robustness against more general and realistic image corruptions. 

\emph{Non-robust features}: In recent work, adversarial examples are shown to carry non-robust features \cite{ilyas2019adversarial} which can be utilized for differentiating the target categories: Interestingly, one can even train a classifier on pure adversarial examples derived from a base classifier, or even with adversarial noise samples (i.e., crafted random-noise samples which the base classifier holds high confidence that they belong to one of the target categories).  

\emph{Other geometric explanations}: 
One limitation of the most widely accepted model-linearity theory is that it disagrees with the experiments showing that linear classifiers do not always suffer from adversarial examples, and adversarial examples that affect deep networks should be different from those for linear classifiers. To support the latter claim, the boundary tilting perspective \cite{tanay2016boundary} states that adversarial examples exist when a decision boundary lies close yet not perfectly aligned to the submanifold of sampled data. Therefore, the perturbed images are likely to cross the boundary. 
In \cite{shamir2019simple}, adversarial examples are considered a natural consequence of the geometry of $\mathds{R}^n$ with the $L_0$ metric. 
Under the assumption that no category is reserved for ``don't know'' and the data distribution is not excessively concentrated, it is further proved that adversarial examples are hard to avoid \cite{shafahi2018adversarial}. 

\section{Hypotheses and Verification} \label{sec:hypotheses_verification}
This section is composed of several subsections, each containing a hypothesis on the causes of adversarial examples and the verification process. The descriptions include the inspiration of the hypothesis, design of the experiment and customized model, as well as results of the controlled experiments. 
Before presenting the major contributions, it is worth reiterating that the focus of the paper is by no means proposing the state-of-the-art attacks or defenses. In addition, we may lose the contrast in the results under strong attacks. 
To provide the reader more clear comparisons as well as illustrating the easiness of attacking the classifiers, we adopt a fast weak and untargeted attack, namely Fast Gradient Sign Method (FGSM) \cite{goodfellow2014explaining}, provided by IBM-ART \cite{art2018}. 

Based on the prediction accuracy and confidence in adversarial examples, we may rank the classifiers into four types in a descending order w.r.t the adversarial robustness:
\begin{enumerate}
    \item robust: high accuracy, high confidence; 
    \item reliable: high accuracy, low confidence; 
    \item unreliable: low accuracy, low confidence; 
    \item misleading: low accuracy, high confidence. 
\end{enumerate}
The degradation in the descending order reflects a gradual shift from being confident of the correct predictions to being obsessed by the incorrect ones. 
In verification processes where critical components (e.g., layers and loss functions) of the model is modified, we evaluate the robustness of the classifier at both the accuracy and confidence levels. 

\subsection{Linearity of the classifier} \label{subsec:linearity}
\begin{hypothesis}
Adversarial perturbations can be magnified by the linear coefficients in the model, which further result in high prediction confidence for misclassification.  \label{hyp:linearity}
\end{hypothesis}

\textbf{Reasoning process}: According to \cite{goodfellow2014explaining}, the output of $w^Tx$ may differ substantially from that of $w^T(x + \Delta x)$. 

\textbf{Design principles}: 
To weaken the linearity of classifiers, we can decrease the scale of the linear coefficients via $L_2$ normalization implemented in weight decay: higher weight decays in the optimizer settings indicate larger $L_2$ penalties. 

\textbf{Technical details}: We train fully-connected neural-network classifiers on MNIST \cite{lecun1998mnist} with varying weight decays in optimizer settings. The common network model consists of 4 layers as 784(relu)-200(relu)-200(relu)-10(softmax), with cross-entropy as the loss function. 
The classifiers are trained for 10 epochs with a batch size of 128 and Adam optimizer with the learning rate of 0.01. 
After obtaining the classifiers, we conduct FGSM with varying strength from and evaluate both the prediction accuracy and confidence on those generated adversarial test samples. 

\begin{figure}[tp]
    \centering
    \smallskip
    \begin{subfigure}[]{0.45\linewidth}
        \centering
        \includegraphics[width=.8\textwidth]{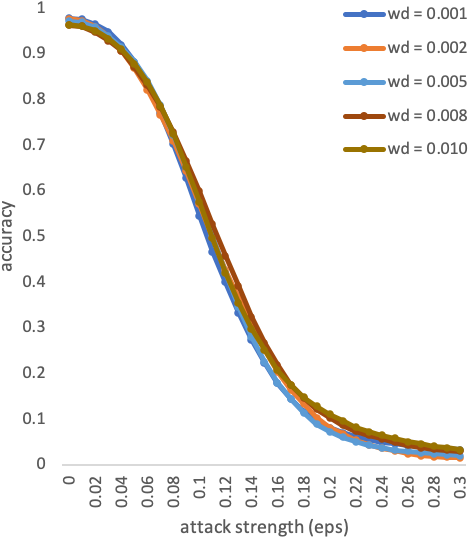}
        \caption{accuracy}
        \label{subfig:wd_accuracy}
    \end{subfigure}
    ~
    \begin{subfigure}[]{0.45\linewidth}
        \centering
        \includegraphics[width=.8\textwidth]{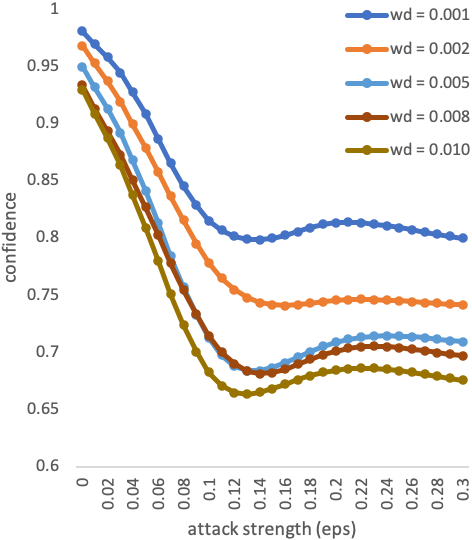}
        \caption{confidence}
        \label{subfig:wd_confidence}
    \end{subfigure}
    \caption{\label{fig:wd_result} Adversarial robustness of classifiers with different linearity: higher weight decay (wd) means stronger $L_2$ normalization and smaller absolute values in linear coefficients. }
\end{figure}

\textbf{Experiment results}:
Fig. \ref{fig:wd_result} plots both average accuracy and prediction confidence for classifiers trained with different weight decays. At the accuracy level (Fig. \ref{subfig:wd_accuracy}), the curves frequently intersect with one another as the attack strength increases. Based on the accuracy curves only, we cannot draw the conclusion that the loss of accuracy on adversarial examples results directly from the linearity of classifiers. 
It is, however, safe to argue that model linearity is related to the high prediction confidence according to the well-separated curves on average confidence (Fig. \ref{subfig:wd_confidence}). 
Moreover, the small bounce near $\epsilon = 0.14$ reflects the shift for most samples from a decrease of probability on the correct category to an increase of probability on the incorrect category. 

\subsection{Categories as events in a probability space} \label{subsec:sum1}
\begin{hypothesis}
Classifiers tend to assign higher confidence at adversarial examples because all output probabilities must add up to 1. \label{hyp:one-sum}
\end{hypothesis}

\textbf{Reasoning process}: Similar arguments have been raised in \cite{fan2018towards} and \cite{shafahi2018adversarial}. Due to the lack of cushion classes for cases such as ``don't know,'' the classifier has to pick the remaining category in high confidence once it rules out all that is impossible. 
Instead of learning the general features and activation, the classifier learns a posterior probability given the input must belong to one of the target categories. 

\textbf{Design principles}: To break the constraint on one-sum probabilities, we replace the softmax activation for the final layer with sigmoid for each neuron, and substitute the cross-entropy (CE) loss with binary cross-entropy (BCE) loss. Consequently, each output probability satisfies its own constraint on belonging to the range of $[0, 1]$ and the sum of probabilities now ranges from 0 to the number of categories. 
Such a configuration is commonly used for training multi-label classifiers. 
For single-label classification problems, the category with highest output probability is selected as the prediction. 

\textbf{Technical details}:
We train a pair of classifiers with the aforementioned two combinations of activation and loss functions, following the exact same configurations as described for verifying the previous hypothesis in Section \ref{subsec:linearity}. 

\begin{figure}
    \centering
    \includegraphics[width=.65\linewidth]{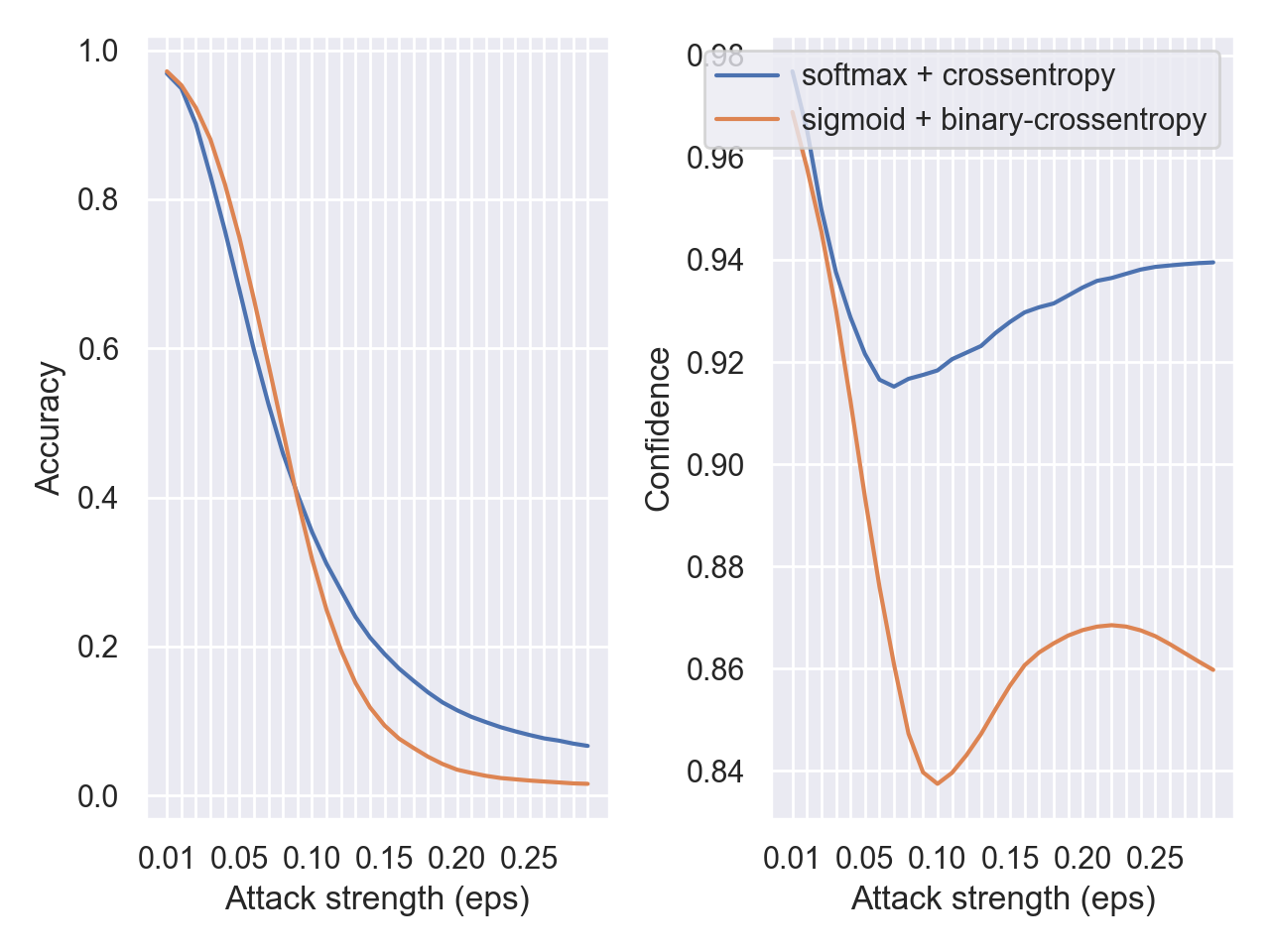}
    \caption{Robustness comparison on classifiers trained with (i.e., softmax + crossentropy) and without (sigmoid + binary-crossentropy) the one-sum constraint on output probabilities. }
    \label{fig:sum1_result}
\end{figure}

\textbf{Experiment results}:
Fig. \ref{fig:sum1_result} reveals interesting behaviors of the two classifiers under adversarial attacks in terms of classification accuracy and prediction confidence. 
One obvious observation is that the prediction confidence becomes lower when the one-sum constraint is lifted, which verifies hypothesis \ref{hyp:one-sum}. 
When the strength of the attack is weak ($\epsilon < .1$), accuracy decreases faster for the classifier trained using softmax and CE. When the attack becomes stronger ($\epsilon > .1$), however, accuracy from the one trained using sigmoid and BCE drops to a lower value. 
One may notice that the turning points for both accuracy and confidence coincide at somewhere around $\epsilon = .1$. Such a flipping point can be explained by the following phases\footnote{Although FGSM is a one-shot attack, we may treat a series of them with increasing $\epsilon$ as an evolution of a single attack for analysis. } of attacking and the one-sum constraint. 

Let us assume that the difficulty varies for adversarially perturbing a sample and alternating its predicted label. Accordingly, a particular attack that drives one sample to phase one may turn another sample to phase two. 
When attacks are weak during the first phase, their major effect is to decrease the probability in the correct category until the output label is changed. Under the one-sum constraint, the probability decrease on one category would lead to an increase of probability on at least another category. On the contrary, the fall-and-rise bond is decoupled by sigmoid and BCE. 
Thus, the one-sum constraint from softmax and CE accelerates the phase of confidence decrease in the correct category.
As attacks become stronger, they further push the prediction confidence on an incorrect category to climb higher at phase two. When the one-sum constraint is lifted by sigmoid and BCE, the prediction confidence for the incorrect class can increase more aggressively without considering other classes, and the accuracy may drop to near 0. In contrast, any further confidence increase under the one-sum constraint with softmax and CE would require a decrease of probability at some other categories. 
Finally, the confidence fall-back for the classifier with sigmoid and BCE after $\epsilon = .2$ reflects that the decrease of probability from the hard-to-attack samples at phase one outweighs the increase of probability from those easy-to-attack samples at phase two. 

\subsection{Combination of linearity and one-sum probabilities} \label{subsec:linearity-probability}
\begin{hypothesis}
Adversarial examples result from the combination of two reasons: (1) linearity of the classifier, and (2) the one-sum requirement on output probabilities. 
\end{hypothesis}

\textbf{Reasoning process}: Experiment results from the previous two hypotheses imply that linearity and one-sum constraint cause high prediction confidence on adversarial examples, but each of them alone may not be sufficient for explaining the significant loss of accuracy under adversarial attacks. This motivates us to consider whether the combination of the two is actually the game-changer.     

\textbf{Design principles}: 
To simultaneously remove the linear coefficient and break the one-sum constraint, we adopt MLP-PNN (denoted by PNN) and Density Estimator (DE), which are proposed in our previous work \cite{li2020continual}.  
Alternatively, one may follow a combined approach by introducing weight decay as in verifying Hypothesis \ref{hyp:linearity} and performing the substitution as we did for Hypothesis \ref{hyp:one-sum}. 
The difference between those two strategies is that the proposed PNN and DE completely remove the linear weights in the final layer (or equivalently setting to them as in unit scale 1), whereas weight decay penalizes large scale of coefficients at all layers.  

\textbf{Technical details}: 
We can view the classifiers in two parts: the feature extractor and the head. Two options are available for the feature extractor part: we may either follow the same architecture with fully-connected layers as the one for verifying Hypothesis \ref{hyp:linearity} or add two convolutional layers at the bottom. As for the network head, we experiment with three options: fully-connected layer (FC) with cross-entropy loss, and the two proposed architectures PNN and DE with binary cross-entropy loss. More technical details on PNN and DE are provided in Section \ref{subsec:MPL-PNN}.

\begin{figure}
    \centering
    \smallskip
    \begin{subfigure}[]{0.45\linewidth}
        \centering
        \includegraphics[width=.8\textwidth]{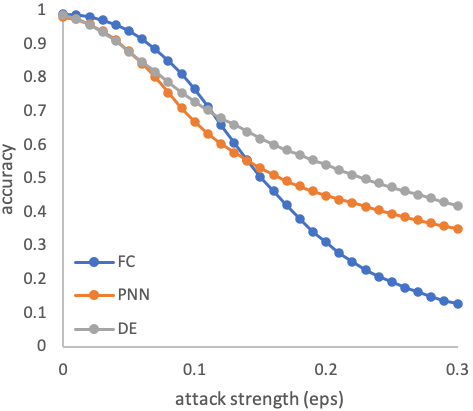}
        \includegraphics[width=.8\textwidth]{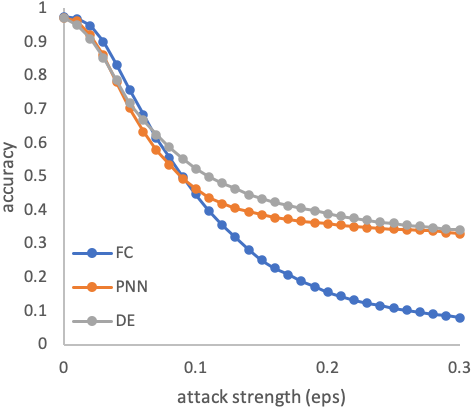}
        \caption{accuracy}
        \label{subfig:head_accuracy}
    \end{subfigure}
    ~
    \begin{subfigure}[]{0.45\linewidth}
        \centering
        \includegraphics[width=.8\textwidth]{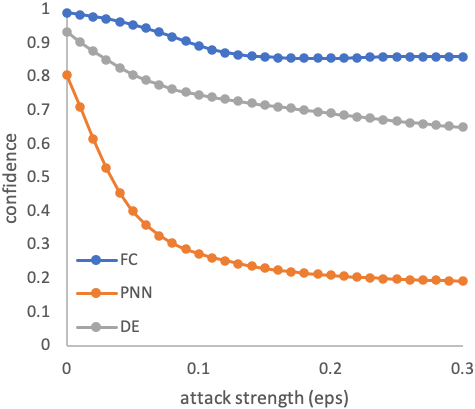}
        \includegraphics[width=.8\textwidth]{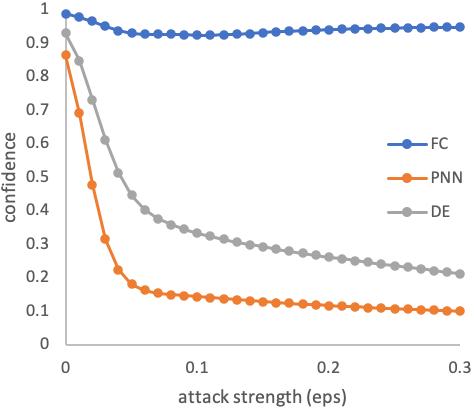}
        \caption{confidence}
        \label{subfig:head_confidence}
    \end{subfigure}
    \caption{\label{fig:fc-pnn-de_result} Adversarial robustness for classifiers with different heads, tested with two types of feature extractor: convolutional (top row) and fully-connected layers (bottom row).  }
\end{figure}

\textbf{Experiment results}:
According to Fig. \ref{fig:fc-pnn-de_result}, the proposed PNN and DE appear to be more adversarially robust than the commonly used MLP (bottom) or CNN (top), which are equipped with heads of fully-connected (FC) layers. 
Similar to the analysis for Fig. \ref{fig:sum1_result} in the previous hypothesis, the decrease of accuracy can be roughly divided into two phases (i.e., probability decrease of the correct class and probability increase of the incorrect class, for FC in particular). 
As the strength of attacks increases, the classification accuracy from PNN and DE gradually converges at a significantly higher level than FC; meanwhile, the prediction confidence keeps decreasing. 
If we recall the four types of classifiers listed at the beginning of Section \ref{sec:hypotheses_verification}, MLP and CNN should be considered as misleading because of their low classification accuracy and high prediction confidence. 
In contrast, the proposed PNN and DE are at least one level higher: being somewhat unreliable yet sensitive to perturbations in terms of confidence drop. 
Overall, results from Fig. \ref{fig:fc-pnn-de_result} demonstrate that the combination of linearity and one-sum constraint brings a stronger impact to the existence of adversarial examples.  

The comparison between PNN and DE reflects a trade-off from the flexibility of bias terms in the head layer. One major difference between PNN and DE is that the bias terms in the final layer are predefined and fixed in PNN but trainable in DE. 
As a consequence, DE achieves slightly higher accuracy on adversarial examples yet in much higher confidence. We will leave the interpretation and validation of such an observation as future work.  

\subsection{Path-connected regions from classifiers} \label{subsec:path-connect}
\begin{hypothesis}
Adversarial examples exist at uncertain ``bridges'' which are created by the classifier for connecting samples of the same category in a path-connected manner.  \label{hyp:bridges}
\end{hypothesis}

\textbf{Reasoning process}: As pointed out by \cite{fawzi2018empirical}, neural-network classifiers tend to partition the input space into path-connected regions. Given that classifiers have limited finite capability of approximating the ground-truth partition, there will often be samples that are consistently misclassified during training, especially in early epochs. Henceforth, we will refer to those as illusive samples. To connect those separated samples with path-connected regions, a network may need to deform its decision boundaries, creating somewhat arbitrary bridges that pass through uncertain regions in between. 
Our guess is that if we exclude those illusive samples during training, the obtained classifier may become more robust because less uncertain areas are required to establish the connection. 

\textbf{Design principles}:
We train classifiers with the same model and configurations on the following different training sets: 
\begin{itemize}
    \item the entire original training set (control);
    \item original training set w/o illusive samples (experimental);
    \item original training set after randomly removing the same number of samples as the number of illusive samples in the experimental group (control).
\end{itemize}
To identify illusive samples, we train multiple shallow classifiers and compute training statistics such as counting the times that a particular training sample has been correctly classified. Shallow classifiers are preferred for this procedure because deep ones with huge model capacity may easily classify most (if not all) training samples correctly. 

\textbf{Technical details}:
The illusive samples are selected based on the training statistics obtained from training 10 randomly-initialized shallow classifiers\footnote{\url{ https://keras.io/examples/cifar10_cnn/}} on CIFAR10 \cite{krizhevsky2009learning} for 25 epochs. If a training sample is always misclassified by all of the 10 classifiers, we consider it as an illusive sample. 
We then train a deeper classifier on CIFAR10 using the CleverHans toolbox \cite{papernot2016technical} on the three datasets listed above. 

\begin{figure}
    \centering
    \smallskip
    \includegraphics[width=.72\linewidth]{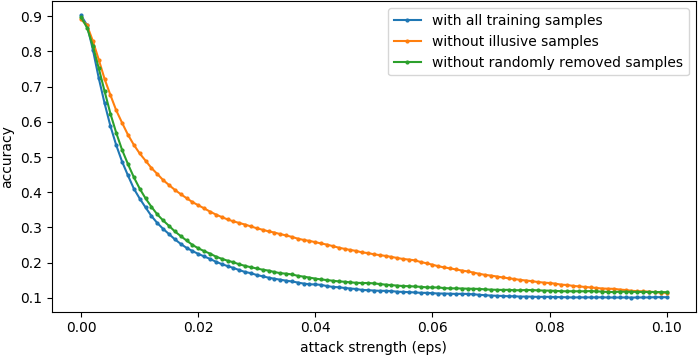}
    \caption{Training without consistently-misclassified illusive samples help enhance robustness against adversarial attacks.}
    \label{fig:illusive-sample_result}
\end{figure}

\textbf{Experiment results}: 
Fig. \ref{fig:illusive-sample_result} shows the accuracy on adversarial test samples for the three classifiers trained with different training sets. When no defense is performed, training without illusive samples leads to higher accuracy on adversarial examples with a slight accuracy drop on clean test samples. Therefore, the adversarial robustness of a classifier is related to the illusive (i.e., hard) samples in the training set. 

\textbf{Self-reflection}: 
It should be noted that to fully justify the hypothesis rigorously is extremely challenging: one shall define, locate, or even visualize the uncertain ``bridges'' in an exceedingly high-dimensional image space. Our experiment, alternatively, presents an indirect verification via results that are consistent with hypothesis \ref{hyp:bridges}. 
Another clarification is that path-connectedness is not a negative characteristic at all time: it fulfills generalization to unseen test samples. It is, however, more desirable should the classifier be capable of preserving only connections that are necessary.  
We have attempted to allow multiple clusters and distributions for each category so that unnecessary ``bridges'' can be avoided in the partition. Unfortunately, almost all samples in a category are assigned to the same cluster by the classifier. 
A key part of future studies is to break the constraint on path-connected regions.  
\subsection{Excessive number of target categories} \label{subsec:excessive-classes}
\begin{hypothesis}
Classifiers trained for fewer target categories tend to be more robust than those trained for more target categories. \label{hyp:classes}
\end{hypothesis}

\textbf{Reasoning process}: As neural networks tend to classify the input space into path-connected regions \cite{fawzi2018empirical}, the categories are expected to intertwine with each other as the total number increases. In addition, given that classifiers tend to place all input samples (even randomly sampled in the input space) close to the boundaries \cite{mickisch2020understanding}, more categories would give rise to more possibilities for attacks.  

\textbf{Design principles}: 
For demonstrating the impact from the number of target categories, we train 9 classifiers (offline) on subsets of categories from MNIST \cite{lecun1998mnist} by gradually adding the categories in a sequential order (i.e., additive mode), starting from the first two categories $\{0, 1\}$ to all 10 categories $\{0, 1, \cdots, 9\}$. 
To rule out the potentially dominant impact from the change of the total number of training samples, we also train those 9 classifiers by keeping (1) the total number of training samples constant and (2) number of samples for each category balanced (i.e., constant mode). 
For all classifiers, the number of neurons in the last layer is equal to the number of categories on which they are trained. 
It is desirable to tune the initial accuracy on clean test samples (i.e., $\epsilon = 0$) so that they are at a similar level for all classifiers trained on various number of target categories, thus the accuracy drops under attack shall be ascribed to the loss of robustness. One reason for using MNIST in our experiment is that such a goal is much easier to reach using this dataset.\footnote{MNIST is perhaps one of the few real datasets on which a modern classifier can achieve similar accuracy as on its subsets of categories.} For most public datasets, adding target categories will decrease the accuracy as the task becomes noticeably harder.  
We then evaluate the robustness of those classifiers using adversarial examples from FGSM attack at varying strength. When evaluating a particular classifier, test samples from only the trained categories are involved.  

\textbf{Technical details}: 
We adopt the model architecture from the official MNIST example from PyTorch,\footnote{\url{https://github.com/pytorch/examples/tree/master/mnist}} and adjust the output shape of the last layer to match the number of categories. 
Following the design principles, we train two groups of classifiers in both the additive and constant mode. For the constant mode, we fix the number of total training samples to 10,000. 
Similar to the previous verification processes, the robustness of the classifiers are examined via attacks with various strength. 

\begin{figure}
    \centering
    \smallskip
    \begin{subfigure}[b]{0.48\linewidth}
        \centering
        \includegraphics[width=.9\linewidth, height=3.6cm]{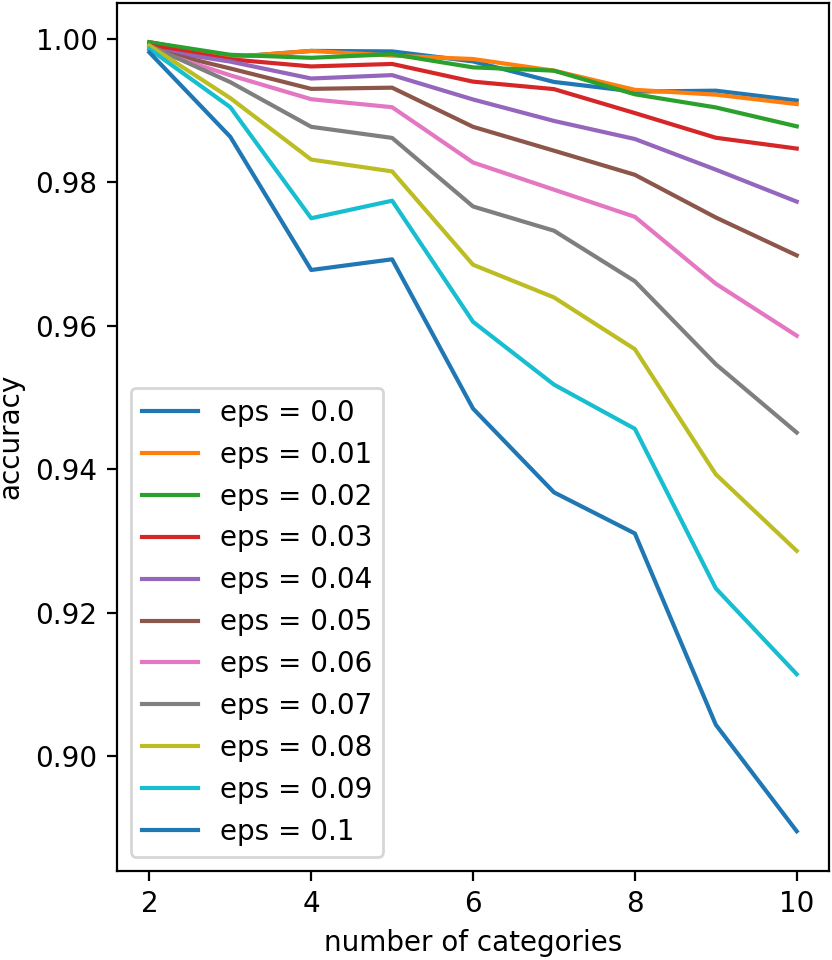}
        \caption{\centering additive mode: including all available training samples}
        \label{subfig:number-of-class_additive}
    \end{subfigure}
    ~
    \begin{subfigure}[b]{0.48\linewidth}
        \centering
        \includegraphics[width=.9\linewidth, height=3.6cm]{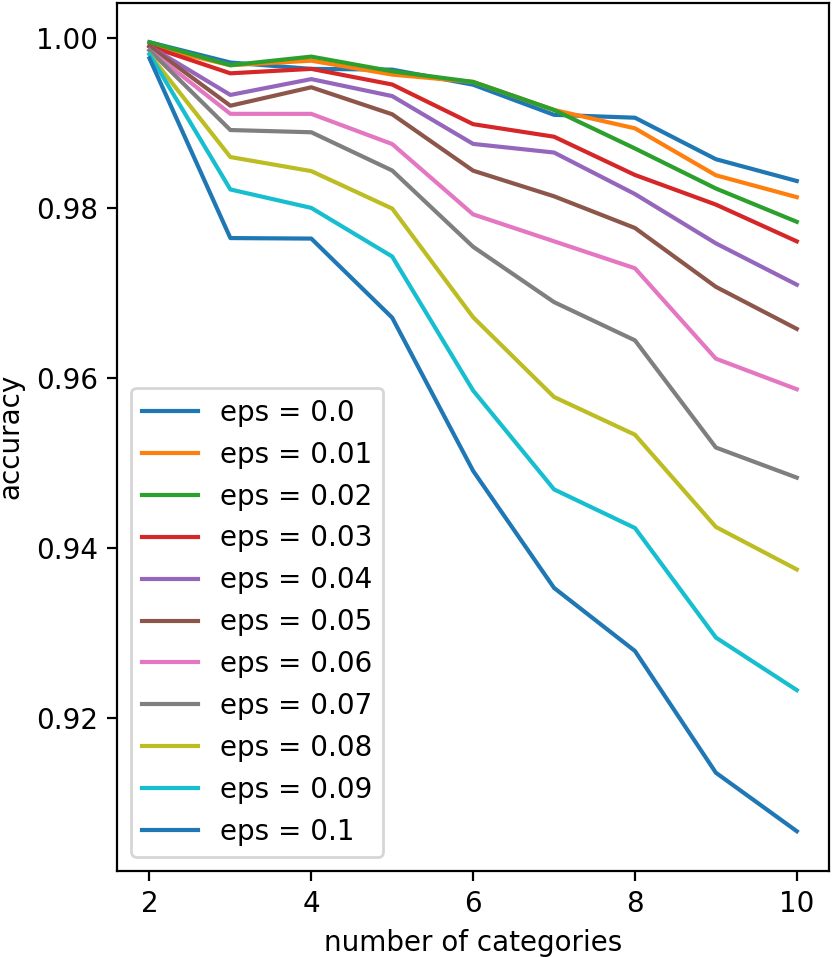}
        \caption{\centering constant mode: 10,000 training samples (balanced)}
        \label{subfig:number-of-class_constant}
    \end{subfigure}
    
    \caption{Robustness of classifiers decreases as the number of target categories increases. $\epsilon$ (eps): the strength of attacks. }
    \label{fig:number-of-class_result}
\end{figure}

\textbf{Experiment results}:
Fig. \ref{fig:number-of-class_result} illustrates the loss of robustness as the number of target categories increases, regardless of the number of total training samples.  
Similar trend is also observed from classifiers with PNN and DE, the two architectures introduced in the previous subsection (and details in Section \ref{sec:support}). 

\subsection{Geometry of input spaces} \label{subsec:geometry}
\begin{hypothesis}
The adversarial robustness of a classifier depends on the geometry of the input space (i.e., entropy of the distribution of categories). Tentatively, the robustness tends to be positively correlated to the ratio of inter-class distance $d_\text{inter}$ to intra-class distance $d_\text{intra}$ among samples. 
\end{hypothesis}

\begin{figure}
    \centering
    \smallskip
    \begin{subfigure}[t]{0.30\linewidth}
        \centering
        \includegraphics[width=.8\textwidth]{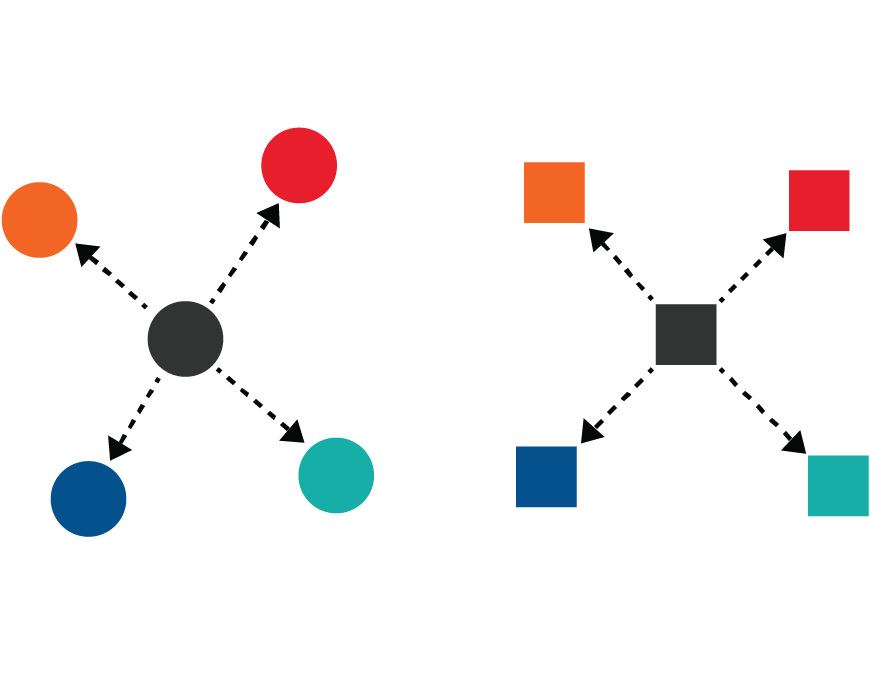}
        \caption{style classes}
        \label{subfig:style_partition}
    \end{subfigure}
    ~
    \begin{subfigure}[t]{0.30\linewidth}
        \centering
        \includegraphics[width=.8\textwidth]{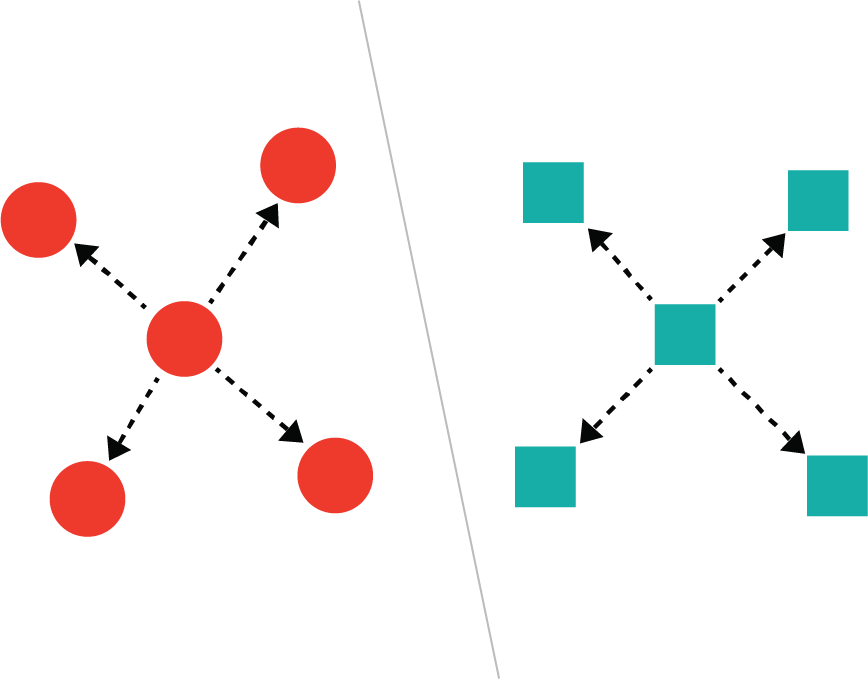}
        \caption{content classes}
        \label{subfig:content_partition}
    \end{subfigure}
    ~
    \begin{subfigure}[t]{0.30\linewidth}
        \centering
        \includegraphics[width=.60\textwidth]{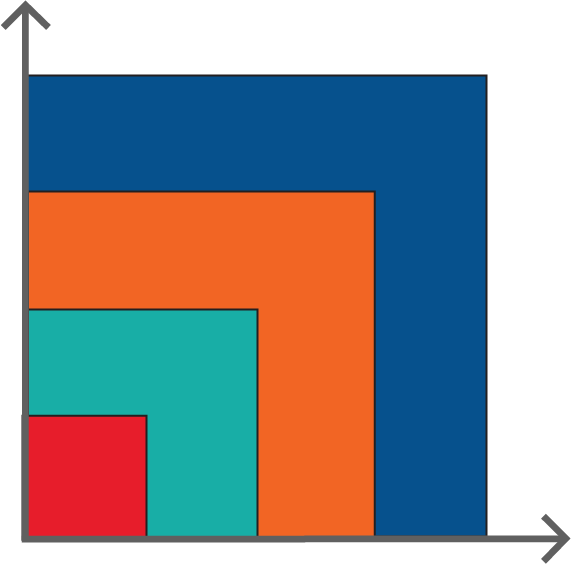}
        \caption{band classes}
        \label{subfig:band_partition}
    \end{subfigure}
    \caption{\label{fig:space_partition}Partition of an input space with different categories}
\end{figure}

\textbf{Design principles}: To verify the hypothesis, we prepare three datasets collected from the same image space (i.e., resized to identical input dimension) with same number of categories and samples yet with significantly different geometry. The datasets are constructed by classes which partition the input spaces in distinctive manners as illustrated in Fig. \ref{fig:space_partition}. 

The style classes (Fig. \ref{subfig:style_partition}) are borrowed from our preset classification benchmark of which the goal is to recognize global photo-editing styles. Starting from a base image (i.e., the dark-gray one in the center) in MIT-Adobe FiveK dataset \cite{fivek}, we obtain 10 stylized version of the same content by applying artistic presets using Adobe Camera Raw, resulting in a total of 11 style categories (including the original ones). 
Due to the control on image content, the intra-class distances $d_\text{intra}$ among samples in the style classes are often larger than the inter-class distances $d_\text{inter}$. 

To match the settings from the style classes, we select 11 categories from the ImageNet dataset \cite{ILSVRC15} and ensure that the initial accuracy on clean test samples are at a similar level as the one from the style classes. In general, a strict partial order may not exist between $d_\text{intra}$ and $d_\text{inter}$ for samples in the content classes, but the ratio of $d_\text{inter} / d_\text{intra}$ is tentatively in between those from the two extreme cases (i.e., style and band classes). 

The band classes are built by equally dividing the input range $[0, 256)$ into 11 bands. For each band class, we generate random samples whose pixel values are independently sampled from the discrete uniform distribution with corresponding range. For instance, the pixel values in the first class are within the range of $[0, 24)$, the second class $[24, 48)$ etc. Under this extreme circumstance, we ensure that $d_\text{intra} < d_\text{inter}$ on average. 

With the above unique settings and everything else controlled, the training process can be viewed as learning to partition the entire input space under the constraints from different sets of ``anchor points'' (i.e., training samples from distinctive distributions). The classifier needs to deform its decision boundary to meet the labels of the training samples, leading to a divergence on the geometry of the learned space.  

\textbf{Technical details}:
Each one of the 11 categories in the three datasets has 1,000 images for training, and 50 images for testing. All classifiers are trained from scratch based on the official ImageNet example code from PyTorch.\footnote{\url{https://github.com/pytorch/examples/tree/master/imagenet}} 
We choose ResNet-50 \cite{he2016deep} with 11 neurons in the final FC layer as the architecture of the model. Except for a smaller batch size of 48, all other hyper-parameters remain the same as the default. The normalization procedure using mean and standard deviation calculated from ImageNet samples has been disabled for consistent range of input domain and fair comparison. 

After training, checkpoints that reach the highest test accuracy at the earliest epoch are passed to the second half of the experiment, in which we examine the adversarial robustness of the classifiers. 
We perform FGSM attack at various strength to the test set and reevaluate the accuracy on those adversarial examples. 

\begin{figure}
    \centering
    \smallskip
    \begin{subfigure}[t]{0.48\linewidth}
        \centering
        \includegraphics[width=.90\textwidth, height=3.8cm]{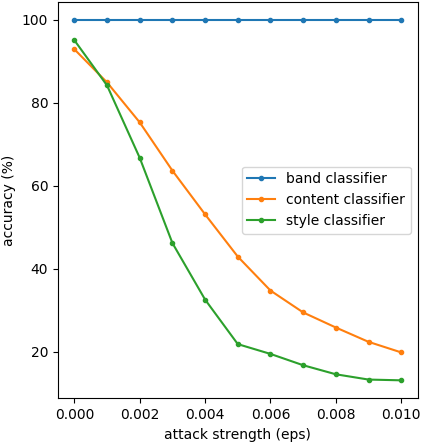}
        \caption{\centering  $\frac{d_\text{inter}}{d_\text{intra}}$: style $<$ content $<$ band}
        \label{subfig:style-content-band_result}
    \end{subfigure}
    ~
    \begin{subfigure}[t]{0.48\linewidth}
        \centering
        \includegraphics[width=.90\textwidth, height=3.8cm]{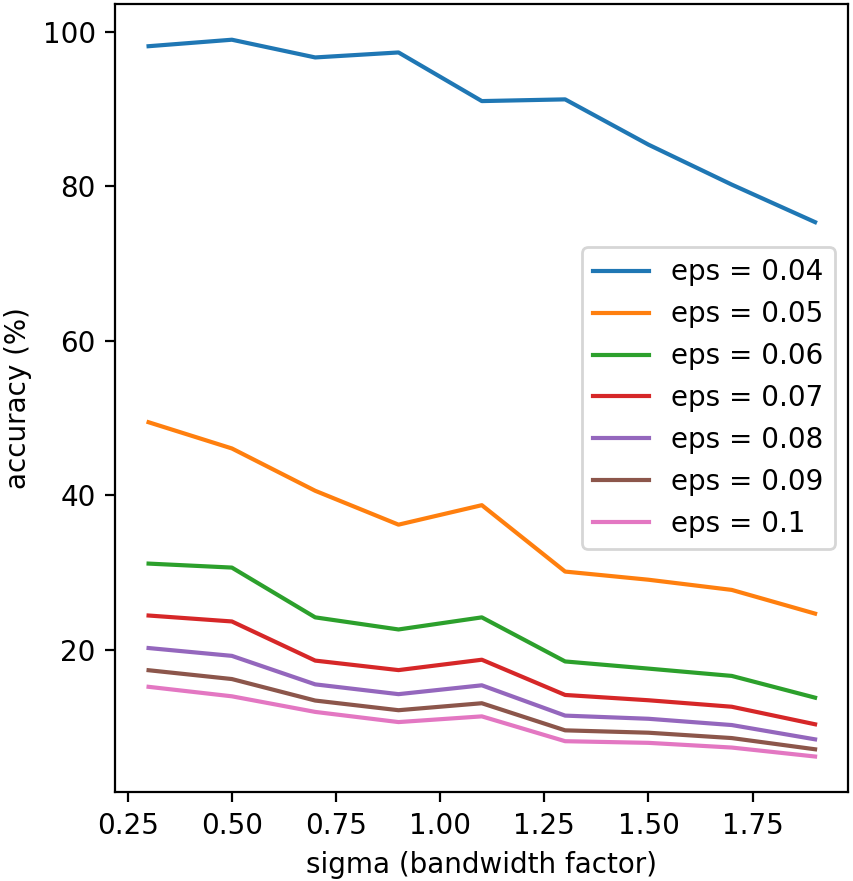}
        \caption{increasing band overlap}
        \label{subfig:bandwidth_result}
    \end{subfigure}
    \caption{Robustness of classifiers when categorizing same number of classes with different geometry. 
    }
    \label{fig:geometry_result}
\end{figure}

\textbf{Experiment results}: Fig. \ref{subfig:style-content-band_result} shows the test accuracy on adversarial examples generated with varying strength for all three classifiers. The robustness of those classifiers agrees with the order of $d_\text{inter} / d_\text{intra}$ from the geometry of the data.
Moreover, as the overlap among those bands increases, the robustness will decrease (Fig. \ref{subfig:bandwidth_result}). 
Classification on the non-overlapping band classes is considered ``intrinsically'' robust because of the linearly-separable distribution of categories.  

\textbf{Self-reflection}: 
Although the band classes in Fig. \ref{subfig:band_partition} may appear to be trivial, the style classes in Fig. \ref{subfig:style_partition} are not. As most studies on adversarial examples are intoxicated by content classification (e.g., with ImageNet dataset), it is meaningful to consider an essentially distinctive classification problem in which the distribution of categories and geometry of decision boundaries are significantly different. 
In addition, to fully justify the influence of a certain factor, both positive and \emph{negative} variations are required. The negative variation, however, is often missing in previous studies on adversarial defenses. Often when a defense is proposed, it is claimed to increase the robustness by mitigating a specific issue. The increase of accuracy on adversarial examples is sufficient to demonstrate the efficacy of a proposed defense. However, it may not be adequate to prove that the specified issue is indeed the determinant because we only observe the positive variation brought by the defense. To fully demonstrate the role of a claimed factor, the decrease of robustness is required when the issue is exacerbated. In the above verification process, we provide both positive (i.e., the band classes) and negative (i.e., the style classes) variations of the base scenario (i.e., content classes). 
Finally, a more rigorous way to create datasets for proving the hypothesis is to build a generator which can produce or evolve non-trivial data at a specified ratio of $d_\text{inter} / d_\text{intra}$, and we will leave it for future work.

\section{Theory for Model Design} \label{sec:support}
This section further elaborates the theoretical foundation that supports model design in the verifying the hypotheses.

\subsection{Probabilistic Neural Networks (PNNs)}
\begin{figure}
    \centering
    \smallskip
    \includegraphics[width=.75\linewidth]{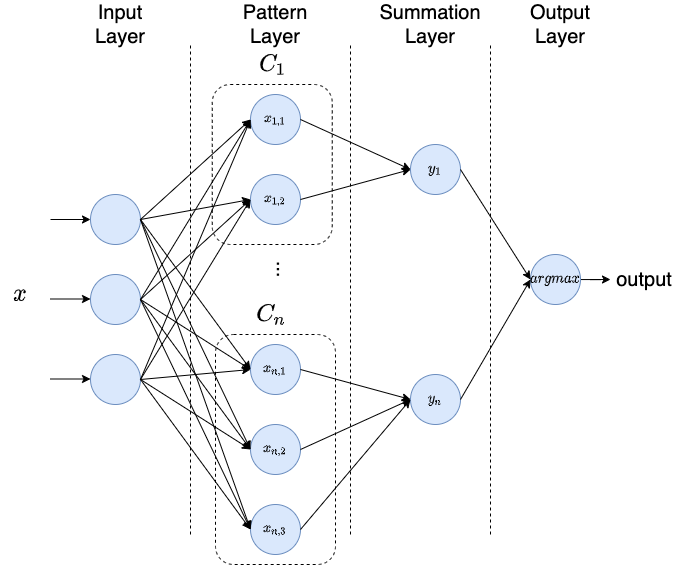}
    \caption{Schematic diagram of probabilistic neural networks. 
    }
    \label{fig:pnn_architecture}
\end{figure}

Probabilistic Neural Networks (PNNs) \cite{specht1988probabilistic} are groups of artificial neural networks widely used in classification and pattern recognition \cite{wu2007leaf, bankert1994cloud}. The PNNs devise probabilistic density function estimators by Parzen window \cite{parzen1962estimation} and then estimate the posterior probability by Bayes' rules \cite{mohebali2020probabilistic}. 
The four-layer architecture of PNN \cite{specht1990probabilistic} is shown in Fig. \ref{fig:pnn_architecture}. The input layer takes the data and passes them to the pattern layer, which performs the density estimation. Each node in the summation layer collects the outputs from the corresponding nodes for each class and produces the probability. The output layer makes a prediction according to the maximum probability from the summation layer. A PNN consists of $n$ sub-networks, each of which is a Parzen-window probabilistic density estimator for a particular class $C_i$. 
Assuming there are $m$ nodes in the $i^\text{th}$ class $C_i$, we can compute the similarity (i.e., prior probability) of a sample $\textbf{x}$ with respect to $\textbf{x}_{i,k}$ (i.e., the $k^\text{th}$ node in the $i^\text{th}$ class $C_i$) as follows \cite{serpen1997performance}: 
\begin{equation}
\label{eq:potential_function}
    K(\textbf{x}, \textbf{x}_{i,k}) = \exp[-\frac{1}{2}(\textbf{x} - \textbf{x}_{i,k})^T \Sigma^{-1}(\textbf{x}-\textbf{x}_{i,k})]
\end{equation}
The posterior probability of \textbf{x} $\in C_i$ is given by
\begin{equation}
\label{eq:summation}
P_i = P(C_i|\textbf{x}) = \phi(K(\textbf{x}, \textbf{x}_{i,1}), \dots, K(\textbf{x}, \textbf{x}_{i,m}))
\end{equation}
in which $\phi(\cdot)$ is a function (e.g., average function) in the summation layer for calculating the posterior probability. 

Different from Multi-Layer Perceptrons (MLP), PNNs output the probabilities for classes according to existing data patterns calculated in the pattern layer. The number of nodes required in the pattern layer, however, may grow exponentially along with the number of training samples. 
As the prediction relies heavily on those data patterns, the process can become complex if there are massive nodes in the pattern layer. Moreover, the error buildup can be significant in terms of both accuracy and generalization.  
Despite the success on large-scale datasets from Long-Sort-Term-Memory (LSTM) \cite{hochreiter1997long} and Convolutional Neural Networks (CNNs) \cite{krizhevsky2012imagenet}, those models fail to estimate an uncertainty \cite{pearce2018uncertainty} based on which one may detect adversarial examples. 
Existing approaches on uncertainty estimation leverage ensemble methods \cite{pearce2018uncertainty} or Monte Carlo sampling procedures \cite{blundell2015weight}. Compared to PNNs, those methods are computationally more expensive and less efficient. 

\subsection{Combination of MLP and PNN} \label{subsec:MPL-PNN}
In \cite{li2020continual}, we combined the PNN and MLP to efficiently produce the prediction confidence. The combined MLP-PNN model is trained with back-propagation to avoid the massive growth of nodes in the pattern layer, and it improves the classification accuracy significantly. The core idea for building the MLP-PNN is to insert hidden layers before the pattern layers so that the number of nodes in the pattern layers can be pre-defined. Therefore, the MLP-PNN model contains five components from bottom to top: an input layer, multiple hidden layers, a pattern layer, a summation layer, and an output layer. 
Let $f(\cdot)$ denote the forward pass provided by the hidden layers, we can then compute the probability $P_i$ for class $C_i$ by substituting $\textbf{x}$ with $f(\textbf{x})$ and defining the $\phi(\cdot)$ in equation (\ref{eq:summation}) as follows. 
\begin{equation}
\label{eq:mlp_pnn_prob}
    P_i = \frac{\sum_{k=1}^m K(f(\textbf{x}), \textbf{x}_{i,k})}{\sum\limits_{k=1}^m K(f(\textbf{x}), \textbf{x}_{i,k}) + m - \max\limits_{\small{k = 1\cdots m}} \{K(f(\textbf{x}), \textbf{x}_{i,k})\}}
\end{equation}

The probability in equation (\ref{eq:mlp_pnn_prob}) can be regarded as the representation of the similarity between $f(\textbf{x)}$ and the nodes of the pattern layer in the $i^\text{th}$ class. Hence, the sum of the output probabilities $\sum_{i=1}^n P_i$ is not necessarily 1. 

Due to the high dimension of $\textbf{x}_k$, calculating the covariance in matrix in equation (\ref{eq:potential_function}) could be intractable in neural networks . Therefore, we assume that the nodes in the pattern layers are independent to each other and the diagonal in the covariance matrix is a pre-defined constant \cite{li2020continual}, denoted by $\sigma$ in equation (\ref{eq:pnn_potention}). We refer to a MLP-PNN with the following potential function in equation (\ref{eq:pnn_potention}) 
as PNN in Section \ref{subsec:linearity-probability}. 
\begin{equation}
\label{eq:pnn_potention}
    K(f(\textbf{x}), \textbf{x}_k) = \exp[-\frac{1}{2}\frac{||f(\textbf{x}) - \textbf{x}_k||}{2\boldmath{\sigma^2}}]
\end{equation}

For simplicity, we further assume the feature vectors (i.e., $f(\textbf{x})$) are independent and calculate their joint probability. The adopted potential function is shown in equation (\ref{eq:de_potention}), where $b$ is the dimension of $\textbf{x}$, $\textbf{x}^{(a)}$ is the $a^\text{th}$ element in $\textbf{x}$, and $\sigma$ is a trainable vector with the same dimension as $\textbf{x}$. We refer to the MLP-PNN with the following potential function in equation (\ref{eq:de_potention}) to Density Estimator (DE) in Section \ref{subsec:linearity-probability}.

\begin{equation}
\label{eq:de_potention}
    K(f(\textbf{x}), \textbf{x}_k) = \prod_{a=1}^b\exp[-\frac{1}{2}\frac{f(\textbf{x})^{(a)} - \textbf{x}_k^{(a)}}{2\boldmath{(\sigma^{(a)})^2}}]
\end{equation}

\subsection{Differences between MLPs and PNNs}

\textbf{Weight Matrix}: Fully-connected layers in MLPs carry the input to a linear function $f(\textbf{x}) = A\textbf{x} + b$. In PNNs, however, there is no weight multiplication for the inputs in the potential functions, as shown in equations (\ref{eq:pnn_potention}) and (\ref{eq:de_potention}).

\textbf{Bias}: Each fully-connected layer contains only one bias vector. On the contrary, the collection of data patterns in the PNNs can be viewed as multiple biases. In conventional PNNs, the data patterns can be learned incrementally. In the proposed MLP-PNNs, the number of data patterns are pre-defined. 

\textbf{Activation function}: For single-label classifiers, the activation function associated with the last fully-connected layer in MLPs is the softmax. As for PNNs, the activation function is in a Gaussian form. 

\textbf{Summation layer}: Summation layers do not exist in the MLPs whereas they are important in PNNs. As we mentioned above, there are multiple data patterns in the PNNs. The output of the pattern layers contains multiple values corresponding to the probabilities calculated from each data pattern for the classes. The summation layer processes the probabilities under certain rules and produces the probability of each class.

\section{Conclusion} \label{sec:conclusion}
In this paper, we verify or partially verify the following causes of adversarial examples: linearity of the model, one-sum constraint on output probabilities, path-connected classification regions, excessive number of target categories, geometry of the input space (i.e., category distribution). We believe this paper will inspire more studies on investigating the root causes of adversarial examples, which in turn provide useful guidance on designing more robust models. Our future works include investigating the probabilistic neural network and dynamic estimator, investigating the influence of the ratio of inter-class and intra-class on the geometry of the input space and break the constraint on path-connected regions.


\bibliographystyle{./bibliography/IEEEtran}
\small{\bibliography{references}}


\end{document}